# Slow Feature Analysis for Human Action Recognition

Zhang Zhang, *Member*, *IEEE*, and Dacheng Tao, *Member*, *IEEE*

**Abstract**—Slow Feature Analysis (SFA) extracts slowly varying features from a quickly varying input signal [1]. It has been successfully applied to modeling the visual receptive fields of the cortical neurons. Sufficient experimental results in neuroscience suggest that the temporal slowness principle is a general learning principle in visual perception. In this paper, we introduce the SFA framework to the problem of human action recognition by incorporating the discriminative information with SFA learning and considering the spatial relationship of body parts. In particular, we consider four kinds of SFA learning strategies, including the original unsupervised SFA (U-SFA), the supervised SFA (S-SFA), the discriminative SFA (D-SFA), and the spatial discriminative SFA (SD-SFA), to extract slow feature functions from a large amount of training cuboids which are obtained by random sampling in motion boundaries. Afterward, to represent action sequences, the squared first order temporal derivatives are accumulated over all transformed cuboids into one feature vector, which is termed the Accumulated Squared Derivative (ASD) feature. The ASD feature encodes the statistical distribution of slow features in an action sequence. Finally, a linear support vector machine (SVM) is trained to classify actions represented by ASD features. We conduct extensive experiments, including two sets of control experiments, two sets of large scale experiments on the KTH and Weizmann databases, and two sets of experiments on the CASIA and UT-interaction databases, to demonstrate the effectiveness of SFA for human action recognition. Experimental results suggest that the SFA-based approach 1) is able to extract useful motion patterns and improves the recognition performance, 2) requires less intermediate processing steps but achieves comparable or even better performance, and 3) has good potential to recognize complex multiperson activities.

✦

## 1 INTRODUCTION

HUMAN action recognition has a wide application prospect in video surveillance, video content analysis, and human computer interactions [5]. As one of the most active topics in computer vision, many works on human action recognition have been reported [12], [13], [14], [15], [16], [17], [18], [19], [20], [21], [22], [23], [24], [25], [26], [27], [28], [29], [30], [31], [32], [33], [34], [35], [36].

### 1.1 Motivation and Overview

In this paper, we propose a group of features to recognize human actions, inspired by the temporal slowness principle, which has been successfully applied to modeling the visual receptive field of the cortical neurons. The temporal slowness principle refers to the primary sensory signal, e.g., the responses of retinal receptors or the gray pixel values of a CCD camera varies quickly within a short period of time; on the other hand, high level responses in a human brain tend to vary slowly for a long time. According to the slowness principle, Wiskott and Sejnowski [1] proposed a nonlinear unsupervised algorithm, i.e., Slow Feature Analysis (SFA), to learn the invariant and slowly varying features from input signals. Berkes and Wiskott [2] employed SFA to learn the self-organized receptive field of cortical neuron from synthetic image sequences. Their experimental results suggest that the learned slow feature functions show many important properties of complex cells in the primary visual cortex (V1), such as the motion direction selectivity and the edge orientation selectivity. Franzius et al. [3] adopted a hierarchical network utilizing both the slowness and the sparseness principles to reproduce the characteristics of place cells, head direction cells, and spatial-view cells. The above findings suggest that the temporal slowness principle or SFA may extract useful motion patterns for human motion analysis.

For human motion analysis, we expect that the SFA learning can discover mapping functions between an input image sequence that varies quickly and the corresponding high-level semantic concepts that vary slowly. Fig. 1 gives an example. Fig. 1a at the left-hand side shows three KTH action sequences [26] that are concatenated together. Fig. 1b in the upper right records the gray values of the three randomly selected pixels "P1," "P2," and "P3" along the time axis. Fig. 1c in the bottom right shows that the action labels at the semantic level change slowly, while the gray values change quickly. Thus, it is necessary to reduce the semantic gap between the quickly varying image signals and the slowly varying action categories.

However, it is impossible to learn a single global function from an entire action sequence because of the complex variations in the high-dimensional image space.

- *Z. Zhang is with the National Laboratory of Pattern Recognition, Institute of Automation, Chinese Academy of Sciences, P.R. China. E-mail: zzhang@nlpr.ia.ac.cn.*
- *D. Tao is with the Centre for Quantum Computation and Intelligent Systems, Faculty of Engineering and Information Technology, University of Technology, Sydney, Australia. E-mail: dacheng.tao@uts.edu.au.*

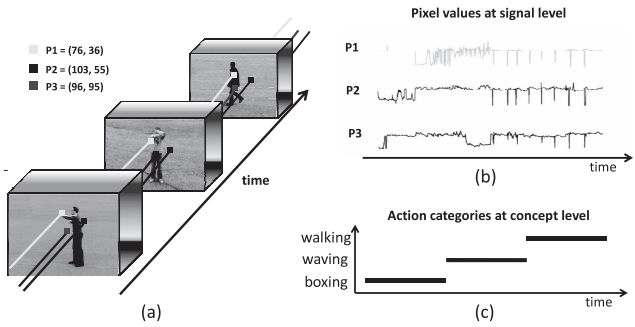

Fig. 1. This example illustrates the relation between slowly varying action concepts and quickly varying pixel values. (a) Three action samples are concatenated into one sequence, including *boxing*, *hand waving*, and *walking*. (b) The gray values of three pixels "*P1*," "*P2*," and "*P3*" over time. (c) High-level representation of action categories over time.

Thus, we propose a local feature-based approach to recognize human actions. The system diagram is shown in Fig. 2. First, a large number of local cuboids are collected by randomly sampling in motion boundaries. Then, a number of slow feature functions are learned from these local cuboids. At the test stage shown in the second row, the detected cuboids in a given test video are transformed by the learned slow feature functions. Afterward, based on the transformed cuboids, we calculate the corresponding feature vector to encode the statistical distribution of slow features. Finally, the support vector machine (SVM) is applied to recognizing human actions in given videos.

### 1.2 Purpose and Contributions

In this paper, the SFA-based features are directly derived from the image sequences for human action recognition. The interesting actions considered in our work include some single person actions such as *walk*, *run*, and *jump*, as well as some multiperson activities, e.g., *meet*, *fight*, and *rob*. Example images are shown in Figs. 8 and 9. These actions can be directly represented by sequential images and do not require the inference of complex spatial-temporal relations.

The main contributions of this paper are summarized as follows:

- SFA has been successfully applied to learning the receptive field of visual cortical neuron [2], [3]. Here, we introduce SFA to the problem of action recognition. To the best of our knowledge, this is the first work that uses the slowness principle or SFA to analyze human motions.
- The original SFA is unsupervised (U-SFA), so the learned slow feature functions ignore the discriminative information. To address this problem for classification task and enhance the selectivity of the learned slow feature functions to different actions, we develop three new SFA learning strategies, which are the supervised SFA (S-SFA), the discriminative SFA (D-SFA) learning, and the spatial discriminative SFA (SD-SFA).
- Instead of using the responses of the learned slow feature functions directly, we propose the Accumulated Squared Derivative (ASD) feature to represent a given action sequence which is a statistical representation of the slow features in an action sequence.
- Extensive experiments on different databases are performed. Besides the widely used KTH [26] and Weizmann data sets [13], two multiple person interaction databases, i.e., the CASIA action data set [43] and the UT-Interaction data set [49], are adopted to validate the advantage of the SFA-based approach to recognize more complex activities.

The remainder of this paper is organized as follows: Section 2 reviews recent works on human action recognition. Section 3 introduces SFA. Section 4 details the SFA-based approach for the recognition of human actions. Data sets used in experiments are introduced in Section 5 and experimental results are presented in Section 6. Section 7 concludes this paper.

## 2 RELATED WORK

In terms of the features used for recognizing human actions, popular methods can be mainly classified into three groups, which are the holistic features-based methods, the local descriptors-based methods, and the biologically inspired methods.

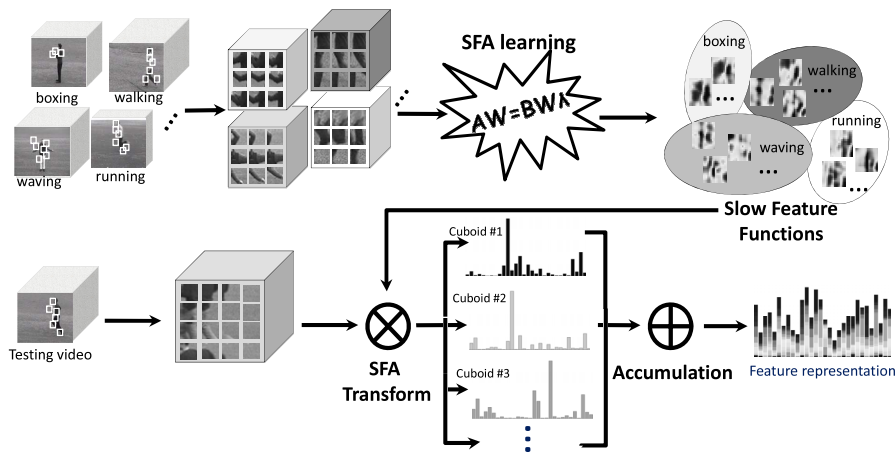

Fig. 2. Diagram of the SFA-based method. First, a large amount of cuboids are collected in training sequences. Then, a number of slow feature functions are obtained by SFA. At the test stage, the detected cuboids in a given video are transformed by the learned slow feature functions. Afterward, a feature vector is calculated to encode the statistical distribution of slow features.

Holistic features utilize some global properties of moving objects (blobs), such as body shape [13], silhouette [14], trajectories of reference joints [31], and motion templates, e.g., motion history image (MHI) and motion energy image (MEI) [15]. The holistic features-based methods require both accurate actor segmentation and perfect body parts tracking. Thus, these methods are sensitive to background motion noises and tracking errors.

Recently, many researchers applied local descriptors to reducing the effects of both background motion noises and tracking errors. The local descriptors-based methods usually include the following intermediate processing steps: interest point detection, local descriptor representation, codebook quantization, and bag-of-words (BoW) representation.

For interest point or salient region detection, a large variety of methods have been proposed. Laptev and Lindeberg [11] extended the 2D spatial interest points to detect 3D spatiotemporal interest points for action representation. The detected local features preserve some translation/rotation invariance, which is desirable for the subsequent action recognition. However, the obtained interest points are too sparse to preserve sufficient information. Dollar et al. [16] presented a detector to obtain a rich number of interest points by using a series of spatiotemporal filters. Oikonomopoulos et al. [17] extended the salient point detector [39] by using the entropy of space-time regions. Recently, Rapantzikos et al. [18] measured the saliency of a space-time region with a global minimization process subject to volumetric constraints. Each of the constraints corresponds to one of the informative visual aspects, such as intensity, color, and motion information. Lee and Chen [19] presented an interest point detector based on the histogram information which can capture large scale structures and distinctive texture patterns. Ke et al. [20] proposed the volumetric features by randomly sampling around one million spatiotemporal volumes with different patch sizes and temporal lengths. Recently, Wang et al. [21] studied different local feature detectors for action recognition and showed that the dense sampling strategy is superior to popular interest point detectors in realistic video settings. However, the dense sampling-based method obtains a large size feature set and thus the subsequent recognition is time consuming.

To describe the collected local interest points or regions, a number of visual descriptors have been proposed. Laptev and Lindeberg [22] proposed and evaluated a series of descriptors including single and multiscale $N$-jets (Gaussian derivatives up to order $N$ along $x, y$, and $t$ axes), the histogram of first order partial derivatives, and the histogram of the optical flow. They showed that the histogram of both spatiotemporal gradients and optic flow significantly performed better than other descriptors. Laptev et al. [23] proposed histograms of gradient orientations (HoG) and histograms of optic flows (HoF) to characterize the local motion and appearance of the space-time neighborhoods of detected interest points. Each space-time neighborhood is divided into a number of cells. For each cell, 4-bin HOG and 5-bin HOF are calculated. Finally, all histograms are concatenated to one feature vector. Scovanner et al. [24] extended the SIFT descriptor to 3D case, so the spatiotemporal information is encoded by a subhistogram. Klaser et al. [25] generalized HoG to the 3D case, where the orientations of 3D gradients in a local region are voted into a set of subhistograms. Then, all subhistograms are concatenated.

After obtaining local descriptors of interest points and salient regions, a codebook of local motion patterns can be obtained by using a clustering algorithm, e.g., K-means. Then, the bag-of-words model [26] can be naturally used to represent an action sequence. To model the co-occurrence relationships among words, a number of topic models, e.g., probabilistic Latent Semantic Analysis (pLSA) [40] and Latent Dirichlet Allocation (LDA) [41], have been introduced to action recognition. Niebles et al. [12] demonstrated the effectiveness of pLSA and LDA for unsupervised learning of action categories. Wong et al. [35] extended pLSA to capture both semantic (content of parts) and structural (connection between parts) information for motion category recognition. Zhang et al. [27] also used pLSA to model the spatial temporal distribution of motion words (MW). Wang and Mori [28] proposed a semilatent topic model to recognize human actions, where each frame corresponds to a "word" instead of a collection of "words" from the space-time interest points. Furthermore, the latent topics directly corresponded to different action categories, so the class labels were exploited in the learning process.

Recently, instead of the bag-of-words paradigm, some other local descriptors-based methods have been proposed to model spatiotemporal relations among local cuboids. Savarese et al. [36] proposed the spatial-temporal correlograms to encode the long range temporal information into the local motion features. Ryoo and Aggarwal [37] proposed a spatiotemporal relationship matching approach for the recognition of multiperson activities, e.g., *push* and *hand-shake*.

On the other hand, Jhuang et al. [32] proposed a biologically inspired system that extended a neurobiological model of motion processing in the visual cortex [42]. This system extracted dense local motion regions by using a set of motion direction sensitive filters. Afterward, the template matching and local maximum operations were conducted alternatively, which were similar to position invariant spatiotemporal feature detectors. By using this biologically inspired feature representation, high recognition accuracies have been reported. Although the proposed SFA-based method is motivated by the studies of the biological vision, it is based on different theories on visual neuron modeling.

## 3 SLOW FEATURE ANALYSIS

Inspired by studies on modeling visual receptive fields of the cortical neurons, many researchers tried to develop algorithms to mimic the functions of visual cortex neurons. Different computational principles and constraints have been proposed to reproduce particular statistical properties of neuron responses to input simulations. Two typical examples are given as follows.

Olshausen and Field [6] proposed an image coding strategy based on the "sparseness structure" principle, according to which an input image is represented by using a small number of descriptors sampled from a large set. The sparseness principle characterizes the simple cells and produces wavelet-like filters to approximate the receptive fields of the simple cells [8]. Recently, the sparseness representation has also been validated to be effective for high-level tasks, e.g., face recognition [44].

The nonnegativity constraint is proposed according to the observations of the behaviors of the primary visual cortex: The firing rates of the simple cells in V1 can never be negative. Hoyer [7] applied this constraint in the sparse coding framework, where both the basis set and the hidden components were constrained to be nonnegative for learning the parts-based representation. Recently, the nonnegative sparse coding has been successfully applied to face recognition [10] and image denoising [9].

However, neither the "sparseness structure" principle [6] nor the nonnegativity constraint [7] models the temporal information in an image sequence. The temporal slowness principle is introduced to model the transformation invariance in a natural image sequence [1]. Sufficient experimental results [2], [3] show that many important properties of visual neurons can be found in the learned slow feature functions. Recently, Franzius et al. [4] proposed a hierarchical model to learn invariant object representation based on the temporal slowness principle. Therefore, SFA is a potential candidate to extract features for action recognition. Mathematically, SFA is defined as follows:

Given an I-dimensional input signal $\mathbf{x}(t) = [x_1(t), \ldots, x_I(t)]^T$ with $t \in [t_0, t_1]$ indicating time, SFA finds out a set of input-output functions $\mathbf{g}(x) = [g_1(x), \ldots, g_J(x)]^T$ so that the J-dimensional output signal $\mathbf{y}(t) = [y_1(t), \ldots, y_J(t)]^T$ with $y_j(t) = g_j(\mathbf{x}(t))$ varies as slowly as possible, i.e., for each $j \in \{1, \ldots, J\}$,

$$\triangle_j = \triangle(y_j) = \langle \dot{y}_j^2 \rangle_t \ is \ minimal, \quad (1)$$

subject to

$$\langle y_j \rangle_t = 0 \ zero \ mean; \quad (2)$$

$$\langle y_j^2 \rangle_t = 1 \ unit \ variance; \quad (3)$$

$$and \quad \forall j' < j : \langle y_{j'} y_j \rangle_t = 0 \ decorrelation, \quad (4)$$

where $\dot{y}$ denotes the operator of computing the first order derivative of $y$ and $\langle y \rangle_t$ is the mean of signal $y$ over time. Equation (1) is the primary objective of minimizing the temporal variation of the output signal, where the temporal variation is measured by the mean of the squared first order derivative. Constraint (2) is presented for convenience only so that Constraint (3) and (4) take a simple form. Constraint (3) means that the transformed signal should carry some information and avoid the trivial solution $y_j(t) = const$. Constraint (4) ensures that different output components carry different types of information and it also induces an order, the first output signal being the slowest one, the second being the second slowest, etc.

If the transformation is linear, i.e., $g_j(\mathbf{x}) = w_j^T \mathbf{x}$, wherein $\mathbf{x}$ is input and $w_j$ is weight, the solution of SFA is equivalent to the generalized eigenvalue problem [1]:

$$AW = BW\Lambda, \quad (5)$$

where $A = \langle \dot{\mathbf{x}}\dot{\mathbf{x}}^T \rangle_t$ is the expectation of the covariance matrix of the temporal first order derivative of the input vector, $B = \langle \mathbf{x}\mathbf{x}^T \rangle_t$ is the expectation of the covariance matrix of the input vector, $\Lambda$ is a diagonal matrix of the generalized eigenvalues, and $W$ is the corresponding generalized eigenvectors. Furthermore, the order of slow features is determined by eigenvalues and where the most slowly varying signal has the lowest index.

The nonlinear transformation can be deemed as the linear transformation in a nonlinear expansion space [1]. The nonlinear expansion function $\mathbf{h}(\mathbf{x})$ is defined by

$$\mathbf{h}(\mathbf{x}) := [h_1(\mathbf{x}), \ldots, h_M(\mathbf{x})]. \quad (6)$$

For example, a quadratic expansion for a 3D input $\mathbf{x} = [x_1, x_2, x_3]$ is $\mathbf{h}(\mathbf{x}) = [x_1^2, x_1 x_2, x_1 x_3, x_2^2, x_2 x_3, x_3^2, x_1, x_2, x_3]$. Afterward, SFA can be performed in the expansion space to obtain nonlinear slow feature functions.

In summary, slow feature functions can be obtained by the following two steps:

- Nonlinear expansion:
  Apply a nonlinear function $\mathbf{h}(\mathbf{x})$ to expand the original signal and centralize $\mathbf{h}(\mathbf{x})$

$$\mathbf{z} := \mathbf{h}(\mathbf{x}) - \mathbf{h}_0, \quad (7)$$

where $\mathbf{h}_0 = <\mathbf{h}(\mathbf{x})>_t$. The centralization makes Constraint (2) valid. In this paper, we use the quadratic expansion, i.e., $\mathbf{h}(\mathbf{x}) = [x_1, \ldots, x_I, x_1 x_1, x_1 x_2, \ldots, x_I x_I]$.

- Solve the generalized eigenvalue problem

$$AW = BW\Lambda, \quad (8)$$

where $A := \langle \dot{\mathbf{z}}\dot{\mathbf{z}}^T \rangle_t$ and $B := \langle \mathbf{z}\mathbf{z}^T \rangle_t$.

Assume the dimensionalities of matrices $A$ and $B$ are $M$, the first $K$ eigenvectors $w_1, \ldots, w_K$ ($K \ll M$) associated with the smallest eigenvalues $\lambda_1 \leq \lambda_2 \leq \cdots \leq \lambda_K$ are the nonlinear slow feature functions $g_1(\mathbf{x}), \ldots, g_K(\mathbf{x})$:

$$g_j(\mathbf{x}) = w_j^T (\mathbf{h}(\mathbf{x}) - \mathbf{h}_0), \quad (9)$$

which satisfies Constraints (2)-(4) and minimizes the objective function (1).

Here, the input-output function computes the output signal instantaneously. Therefore, slow variation of the output signal cannot be achieved by using the temporal low-pass filter, but must be obtained by extracting aspects of the input signal that are inherently slow and useful for a higher level representation.

## 4 SFA-BASED ACTION RECOGNITION

There are four main steps in the SFA-based human action recognition, including *Collection of training cuboids*, *Slow feature function learning*, *Action feature representation*, and *Classification*. In *Slow feature function learning*, we extend the original SFA by using weakly supervised information and spatial information of the training cuboids to obtain discriminative slow feature functions for action classification.

### 4.1 Collection of Training Cuboids

Before the cuboids collection, we perform normalization for each action sequence, so the input signals are of zero mean with unit variance. Then, cuboids can be collected by randomly sampling in informative regions determined by the following two criteria:

- The cuboid is sampled from the foreground region. In the experiments on the Weizmann data set [13] and the KTH data set [26], we directly use the foreground information provided by [13] and [32], respectively. The sizes of the KTH image, the Weizmann image, and the foreground bounding box are $120 \times 160$, $144 \times 180$, and $110 \times 80$, respectively.
- The spatial position $(x, y)$ locates on motion boundaries detected by Sobel operator. The motion boundaries of the frame differences in a bounding box will be returned if its gradient magnitude is larger than a predefined threshold $\delta$. Note for the experiments on CASIA data set [43], we directly detect the motion boundaries from the frame difference without using the foreground bounding box.

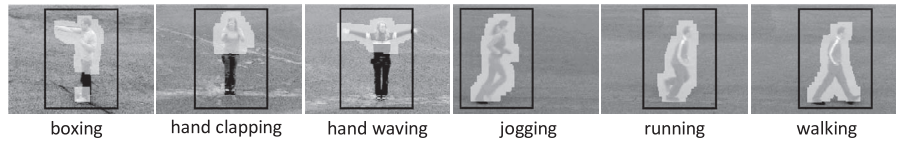

Fig. 3. Examples of training cuboids denoted by the light gray area, where the solid black lines represent the foreground bounding boxes with the size of $110 \times 80$ and the image size is $120 \times 160$.

Initialized at time $t$ and centered at the selected position $(x, y)$, a cuboid is obtained with the size of $h \times w \times d$ ($16 \times 16 \times 7$ in this paper). Fig. 3 presents examples of the collected cuboids. The cuboids in the current frame are indicated by the light gray area. To ensure that most informative regions are selected, the predefined threshold $\delta$ is set as a small value. This setting introduces noise regions in background and shadows. However, the effects of the selected noises can be balanced out by using the statistics over a large number of cuboids.

According to Berkes and Wiskott [2], we reformat each input vector by $\Delta t$ successive frames, so SFA counts the temporal information in the neighbor frames. Fig. 4 shows the reformatting process, where $\Delta t = 3$. After the nonlinear expansion, the dimensionality of input vector increased greatly. For example, the quadratic expansion increases the dimensionality from $n$ to $n + n*(n+1)/2$. Thus, before the nonlinear expansion we perform Principal Component Analysis (PCA) to reduce the dimensionality of the input vector to 50, which is sufficient for the subsequent experiments.

### 4.2 Slow Feature Function Learning

Based on the basic SFA algorithm introduced in Section 3, we investigate four kinds of SFA learning strategies to extract slow feature functions for action recognition, which are the unsupervised SFA (U-SFA), the supervised SFA, the discriminative SFA, and the spatial discriminative SFA.

#### 4.2.1 Unsupervised SFA

The upper left of Fig. 5 shows the U-SFA learning strategy for the extraction of slow feature functions. According to U-SFA, all cuboids collected from different actions are mixed together to learn slow feature functions. Each cuboid is considered as one minisequence. The covariance matrix and the time-derivative covariance matrix are calculated by combining all minisequences. However, the slow feature functions learned by the U-SFA do not encode any supervised information and are universally shared by different actions.

#### 4.2.2 Supervised SFA

The upper right of Fig. 5 presents the S-SFA learning strategy for slow feature function extraction. The collected local cuboids are labeled by action categories. Then, the SFA learning is performed to extract slow feature functions for each action category independently. Finally, the statistical feature is computed with all slow feature functions. However, different actions may share many similar local motion patterns, so different labels to these "common" cuboids are misleading.

#### 4.2.3 Discriminative SFA

To properly introduce the supervised information to the SFA learning, we propose the discriminative SFA shown in the bottom left of Fig. 5. D-SFA is inspired by discriminative sparse coding [45], where a number of sets of discriminative dictionaries are learned, and each set of dictionaries is used to reconstruct a specific image class. Accordingly, D-SFA learns a number of sets of functions and each set of functions is used to slowdown a specific action class.

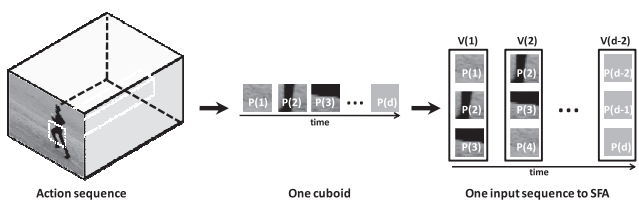

Fig. 4. The reformatting process of the cuboid. The white dashed box is the selected local cuboid in the action sequence. Then, the cuboid is reformatted so that the input vector at each time includes $\Delta t$ successive patches. Here, $\Delta t = 3$.

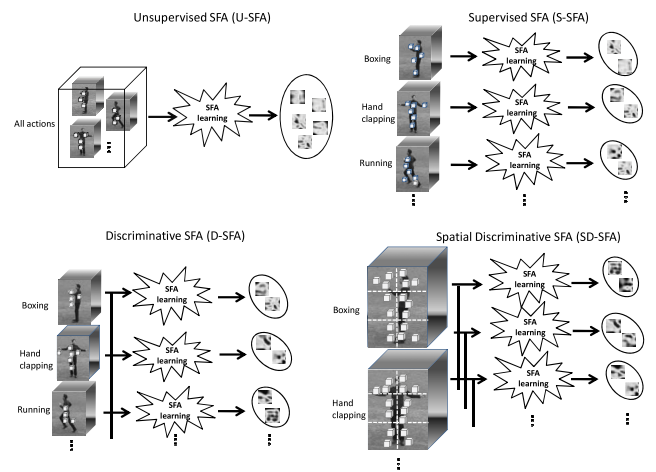

Fig. 5. Four kinds of SFA learning strategies: the unsupervised SFA, the supervised SFA, the discriminative SFA, and the spatial discriminative SFA.

Given $C$ classes of $I$-dims input signals $\{\mathbf{x}_c(t) = [x_{c1}(t), \ldots, x_{cI}(t)] | c \in \{1, \ldots C\}\}$, for the $c$th class, D-SFA finds a set of $J$-dims functions $\mathbf{g}_c(\mathbf{x}) = [g_{c1}(\mathbf{x}), \ldots, g_{cJ}(\mathbf{x})]^T$ to minimize $\Delta(g_{cj}(\mathbf{x}_c)) - \gamma * \Delta(g_{cj}(\mathbf{x}_{c'}))$. Therefore, each learned function makes the intraclass signals $\mathbf{x}_c(t)$ vary slowly, but makes the interclass signals $\mathbf{x}_{c'}(t)$ that are different from class $c$ vary quickly. Assume $\mathbf{g}_c(\mathbf{x}) = [g_{c1}(\mathbf{x}), \ldots, g_{cJ}(\mathbf{x})]^T$ are linear functions, for each $j \in \{1, \ldots, J\}$ D-SFA minimizes

$$\Delta(g_{cj}(\mathbf{x}_c)) - \gamma * \Delta(g_{cj}(\mathbf{x}_{c'}))$$
$$= \langle [g_{cj}(\dot{\mathbf{x}}_c)]^2 \rangle_t - \gamma * \langle [g_{cj}(\dot{\mathbf{x}}_{c'})]^2 \rangle_t \quad (10)$$
$$= w_{cj}^T [\langle \dot{\mathbf{x}}_c \dot{\mathbf{x}}_c^T \rangle_t - \gamma * \langle \dot{\mathbf{x}}_{c'} \dot{\mathbf{x}}_{c'}^T \rangle_t] w_{cj},$$

subject to:

$$\langle g_{cj}(\mathbf{x}_{c \cup c'}) \rangle_t = 0 \quad \textit{zero mean}; \quad (11)$$

$$\langle [g_{cj}(\mathbf{x}_{c \cup c'})]^2 \rangle_t = 1 \quad \textit{unit variance}; \quad (12)$$

$$\forall j' < j : \langle g_{cj'}(\mathbf{x}_{c \cup c'}) g_{cj}(\mathbf{x}_{c \cup c'}) \rangle_t = 0 \quad \textit{decorrelation}, \quad (13)$$

where $w_{cj}$ is the weight vector of the $j$th slow feature function for the class $c$ and $\gamma$ is the tradeoff parameter. D-SFA can be written as a generalized eigenvalue problem:

$$EW = BW\Lambda, \quad (14)$$

where $E = [\langle \dot{\mathbf{x}}_c \dot{\mathbf{x}}_c^T \rangle_t - \gamma * \langle \dot{\mathbf{x}}_{c'} \dot{\mathbf{x}}_{c'}^T \rangle_t]$, $B = \langle \mathbf{x}_{c \cup c'} \mathbf{x}_{c \cup c'}^T \rangle_t$, $\Lambda$ is a diagonal matrix of the generalized eigenvalues, and $W$ is the corresponding generalized eigenvectors.

To obtain nonlinear slow feature functions, we can perform the nonlinear expansion before the D-SFA learning.

### 4.2.4 Spatial Discriminative SFA

The spatial information of motion pixels is useful to infer the body part movements. Furthermore, the body part movements indicate different human actions. For example, the upper region motions might suggest the action of *hand waving*. On the other hand, if there are many frequent motions at the bottom of the foreground box, the action could be *walking* or *running*. The spatial information has been introduced to human action recognition and experimental results suggest its effectiveness. In [27], a polar coordinate centered at the geometric center of the foreground is divided into several areas, and thus the spatial properties of the detected Motion Words can represent relative movements of body parts. In this paper, we divide the foreground bounding box into $I_x \times I_y$ regions. The number of regions along the $x$-axis $I_x$ is 2 because the human body in front view is symmetric. There are three regions along the $y$ axis that correspond to the upper body, the waist and the lower body, respectively. Thus, besides the action category, collected cuboids have their respective region labels. In each region, we extract slow feature functions by D-SFA. Finally, the slow feature functions in all regions are collected together to compute the feature vector for classification.

### 4.3 ASD Feature Representation

In the SFA learning, cuboids are derived from $d$ successive frames. Thus, we compute a statistical feature from $d$ frames to represent an action sequence. According to [33], a very short image sequence is termed an "action snippet." In one snippet, the ASD feature is computed as follows.

First, the motion boundaries in the first frame of the snippet are obtained according to Section 4.1. Then, a number of pixels in the motion boundaries are selected (in this study, for the KTH and Weizmann databases, 25 percent pixels in motion boundaries are selected) as the initial central positions of cuboids with the size of $h \times w \times d$. After the reformation shown in Fig. 4, each cuboid is represented by a vector sequence with the time length of $d - \Delta t + 1$, wherein the vector at each time is obtained by concatenating $\Delta t$ successive patches. With the learned slow feature functions, each input sequence is transformed to a new vector sequence with the size of $K \times (d - \Delta t + 1)$, wherein $K$ is the number of slow functions.

SFA minimizes the average squared derivative, so the fitting degree of a cuboid to a certain slow feature function can be measured by the squared derivative of the transformed cuboid. If the value is small, the cuboid fits the slow feature function. Otherwise, the cuboid does not fit the function. For cuboid $C_i$ and slow function $F_j$, the squared derivative $v_{i,j}$ is

$$v_{i,j} = \frac{1}{d - \Delta t} \sum_{t=1}^{d - \Delta t} \left[ C_i(t+1) \otimes F_j - C_i(t) \otimes F_j \right]^2, \quad (15)$$

where $\otimes$ is the transformation operation.

We then accumulated the squared derivatives over all cuboids to form the ASD feature:

$$f_{ASD} = \sum_i^N V_i, \quad (16)$$

where $N$ is the total number of cuboids in the current snippet and $V_i = <v_{i,1}, v_{i,2}, \ldots, v_{i,K}>^T$.

Since the number of cuboids detected in a snippet may differ from that in another snippet, it is necessary to normalize the feature vector. Here, we perform the $L1$ normalization.

Fig. 6 shows an example of SFA-based feature representation. In this figure, the action example is *jogging* from the KTH data set [26]. In each picture of the first row, the square points are the cuboids collected by random sampling in the motion boundaries. The slow feature functions are obtained by D-SFA. For each action category, the first 200 slow feature functions are preserved to compute the feature vector. Then, the squared first order temporal derivatives are accumulated over all cuboids. For examples, every 55th slow feature function of different action categories is selected and the corresponding squared first order temporal derivative of the first three cuboids and the accumulated value are presented in the middle rows of Fig. 6. Finally, the ASD feature vector is formed by concatenating the accumulated values of all slow feature functions, as shown in the bottom of Fig. 6. We can see that the accumulated values of those slow feature functions corresponding to the action *jogging* are much smaller than the values of slow feature functions obtained from other actions.

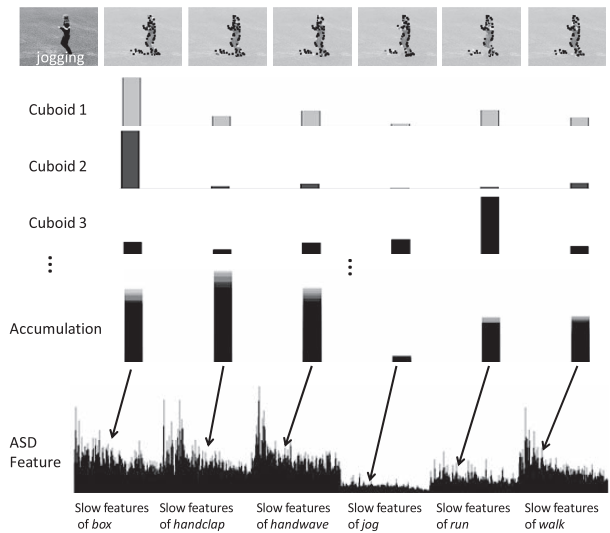

Fig. 6. An example of the computation of the ASD feature. A number of cuboids are collected by random sampling in motion boundaries. Then, the cuboids are transformed by all slow feature functions. Finally, the squared first order derivatives of all cuboids are accumulated to form the feature representation.

For U-SFA, the learned slow feature functions are shared by all action categories. However, the statistical distributions of slow features in different actions are different from each other. Similarly to the bag-of-words model [41], the proposed ASD feature measures the statistical distributions of slow features existing in an action sequence, which can be used for action classification.

For SD-SFA, the feature representation is initially computed in each subregion. Then, the feature vectors in different regions are concatenated into a long feature vector. An example of the SD-SFA-based feature representation is shown in Fig. 7 that is a *running* sample in the KTH data set. The $L1$ normalization is performed on the concatenated vector, so the number of cuboids in different regions is also considered in the whole feature vector.

### 4.4 Classification

After the computation of the ASD feature, we use a linear multiclass SVM [48] for action classification. The dimensionality of the ASD feature is equal to the number of the learned slow feature functions. For example, the D-SFA is performed to obtain 200 slow feature functions for each action category. In the KTH data set, there are six action categories. Thus, for D-SFA, the dimensionality of the ASD feature in the KTH data set is $200 \times 6 = 1{,}200$. The ASD feature is computed from $d$ successive frames. Thus, for a sequence with $N$ frames, we can obtain at most $N - d + 1$ features. Accordingly, $N - d + 1$ labels can be assigned by classifier. Finally, the majority voting rule is used to determine the label of the sequence.

For SD-SFA, the direction of actions should be considered. The actions might occur with inverse direction. To address this problem, we adopt the mirror trick [27], which flips the subregions horizontally. The symmetric feature is used in both training and testing stages.

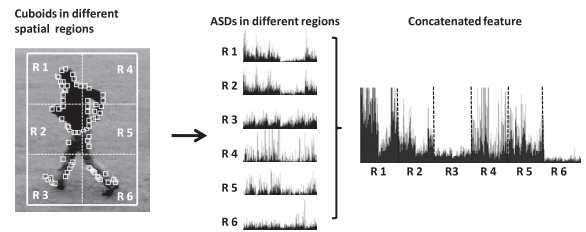

Fig. 7. Example of the SD-SFA-based feature representation. The features in subregions are concatenated into a long vector.

## 5 DATA SETS

To evaluate the capability of the SFA-based method to human action recognition, four action databases, i.e., the KTH action data set [26], the Weizmann human action data set [13], the CASIA action data set [43], and the UT interaction data set [49], are adopted in the experiments.

The KTH data set [26] is the largest action data set (598 action sequences), which contains six types of single person actions (*boxing*, *clapping*, *waving*, *jogging*, *running*, and *walking*), performed by 25 people in four scenarios: outdoors (s1), outdoors with scale variation (s2), outdoors with different clothes (s3), and indoors with lighting variation (s4).

The Weizmann human action data set [13] includes 10 types of single person actions performed by nine subjects, including *bend*, *jump*, *jump jack (jack)*, *jump in place (pjack)*, *run*, *gallop sideways (side)*, *skip*, *walk*, *one-hand wave (wave1)*, and *two hands wave (wave2)*. In total, 90 action sequences are used in our experiments.

The above two data sets have been widely used to evaluate the systems/methods for action recognition. However, they only focus on the recognition of simple single person actions (e.g., running, walking). To validate the advantages of the SFA-based method, we further adopt two multiperson interaction data sets in experiments, i.e., the CASIA action database [43] and the UT interaction database [49].

The CASIA action database [43] contains eight types of single person actions (e.g., *walk* and *run*) and seven types of two-person interactions (*rob*, *fight*, *follow*, *follow and gather*, *meet and part*, *meet and gather*, and *overtake*). In experiments, we chose three interaction categories, i.e., *meet*, *fight*, and *rob*. Here, seven *meet* segments (497 frames), four *fight* segments (451 frames), and four *rob* segments (141 frames) are collected. Each segment is cropped from the original video sequence in time axis, where two people have approached each other closely, so a more subtle visual feature is needed for deciding the type of the interaction. The sample images are shown in Fig. 8, where the size of the image is $240 \times 320$, whereas the foreground area is around $50 \times 35$.

The UT-Interaction data set has been used in the first Contest on Semantic Description of Human Activities (SDHA) [49]. This data set contains action sequences of six interactions: *hug*, *kick*, *point*, *punch*, *push*, and *hand-shake*. Fig. 9 shows some samples in the data set. For classification, 120 video segments cropped based on the ground-truth bounding boxes and time intervals are provided by the data set organizers. These segments are further divided into two

ZHANG AND TAO: SLOW FEATURE ANALYSIS FOR HUMAN ACTION RECOGNITION 443

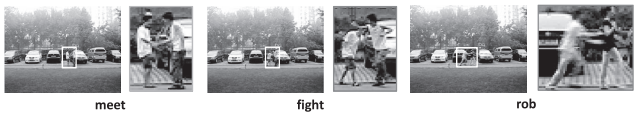

Fig. 8. Sample images of interactions in the CASIA database. There are three categories of two-person interactions, i.e., *meet*, *fight*, and *rob*. For a clear illustration, the foreground areas of the three interactions are enlarged as shown at the right side.

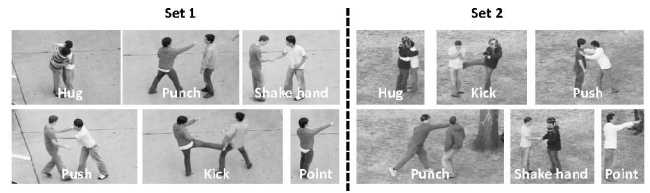

Fig. 9. Example images in the UT-Interaction data set. There are six types of interactions, i.e., *hug*, *kick*, *point*, *punch*, *push*, and *hand-shake*. In total, 120 segments are divided into two sets, where Set #1 is taken at a parking lot with less noise and Set #2 is taken from a lawn on a windy day.

sets (each set has 60 segments and 10 segments per class). Set #1 is taken at a parking lot with less noise and the segments in Set #2 are from a lawn on a windy day. The videos are taken with the resolution of $720 * 480$, 30 fps, and the height of a person is about 200 pixels.

## 6 EXPERIMENTAL RESULTS

In this section, extensive experimental results are reported to demonstrate the effectiveness of the SFA-based approach to human action recognition.

### 6.1 Control Experiment 1: Effects of Different Conditions to Action Recognition

First, we performed the original SFA (U-SFA) on a small data set with different experimental settings. The aim is to evaluate the influence of different conditions to human action recognition. The control experiments were conducted on a subset of the KTH database. The subset includes the first five subjects' action sequences in the outdoor scene *S1*. Here, 3,000 cuboids are collected for the U-SFA learning, where the size of cuboid is $16 \times 16 \times 7$.

Five conditions are considered as follows:

- V1: There are two kinds of inputs, including "Original Frame" and "Frame Difference." The original frame is the gray values of the image pixels. The frame difference is the temporal derivative or equivalently the difference between adjacent frames.
- V2: There are two kinds of representations, including "Holistic Representation" and "Local Representation." For "Holistic Representation," the input vector at each time point is just the values of image pixels in the foreground bounding box with the size of $110 \times 80$. For "Local Representation," an action sequence is represented by a set of local cuboids.
- V3: For "Holistic Representation," two approaches are used for dimension reduction, i.e., "PCA" and "SFA." PCA is performed to reduce the dimensionality from $8,800 \, (110 \times 80)$ to 200. For "SFA," U-SFA is performed to obtain the first 200 slow feature functions.
- V4: For "Local Representation," three strategies for getting cuboids are adopted, i.e., "Grid Sampling," "Motion Boundary in Bounding Box," and "Motion Boundary without Bounding Box." For "Grid Sampling," the sampling points are obtained equidistantly in the foreground bounding box. For the other two sampling strategies, the sampling points are obtained from motion boundaries with or without the help of foreground bounding box.
- V5: For "Local Representation," we test the performance of "Cuboid Classification" and the proposed "ASD Feature." For "Cuboid Classification," the responses to the slow feature functions are directly feed into the SVM classifier. Each cuboid is assigned to a label. The label of one frame is determined by the majority voting rule.

After computing features with the aforementioned different settings, multiclass SVM classifiers [48] are trained for classification. For a given test sequence, the action label is assigned to each frame. And the classification accuracy is given by

$$Acc = \frac{Number\ of\ correct\ classified\ frames}{Total\ number\ of\ frames}. \quad (17)$$

Fig. 10 presents the average accuracies of fivefold cross validation under different experimental settings. Based on the figure, we come up with the following conclusions:

- By using the holistic representation, compared with the PCA-based representation, the SFA is more effective to extract useful features for action recognition. However, in general, the holistic representation cannot perform as well as the local representation.

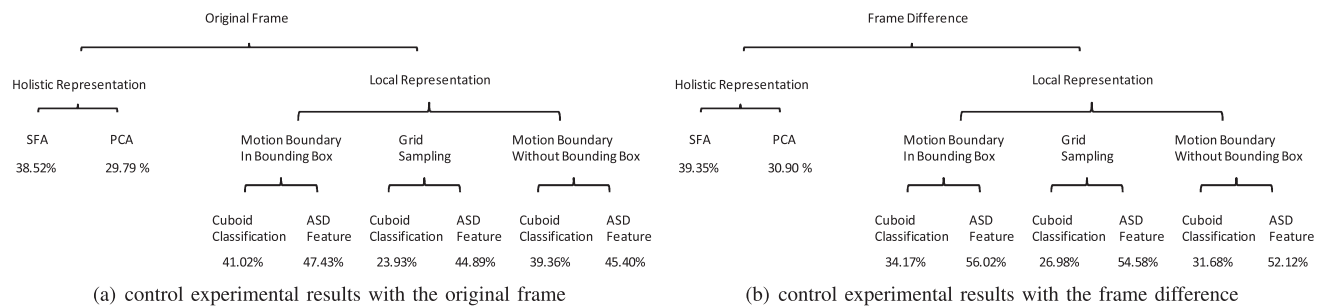

(a) control experimental results with the original frame

(b) control experimental results with the frame difference

Fig. 10. The settings of control experiments and the corresponding experimental results. Each path from the root node to a leaf node denotes one test, where the nodes denote the experimental settings and the under leaf node is the recognition accuracy.

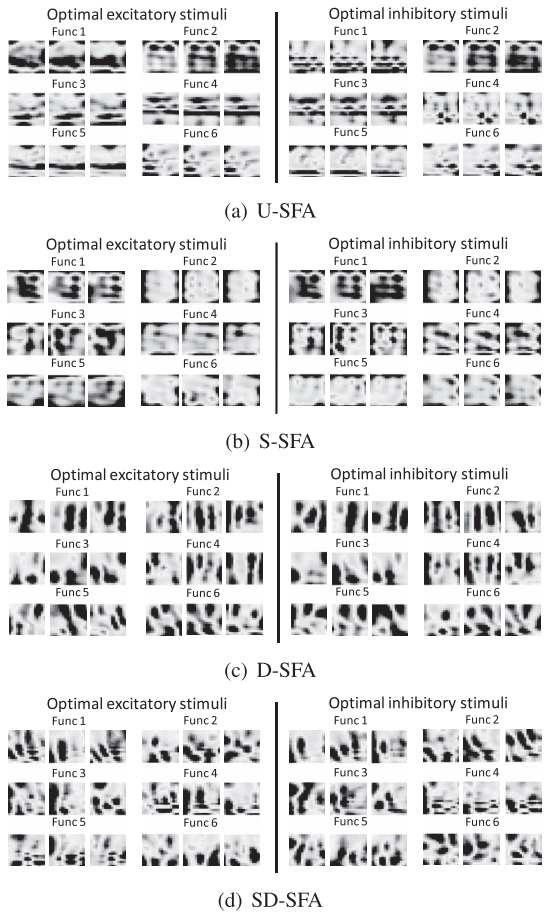

Fig. 11. Some visualizations of the slow feature functions learned by different SFA strategies. (a) The slow feature functions learned by the U-SFA. (b) The functions learned by the S-SFA for the action *boxing*. (c) The functions learned by the D-SFA for the action *boxing*. (d) The functions learned by the SD-SFA for the action *boxing* in the first subregion.

- By using the local representation, the performance of cuboid classification is worse than that of the ASD feature, so the ASD feature is more effective for action recognition. Cuboid classification cannot result in good performance because the supervised information of single cuboid is weak and different actions may share similar cuboids.
- By using the local representation for the ASD feature, the performance of the frame difference is better than that of the original frame. This suggests that the motion information in frame difference is more important for action recognition.
- By using the local representation and the ASD feature, the "Motion boundary in bounding box" achieves the highest accuracy among three sampling strategies, which indicates that the foreground bounding box improves the performance of our methods. However, compared with the other two sampling strategies, the improvement is slight and the proposed SFA-based method does not heavily rely on the foreground bounding box.

In summary, the results of the control experiments suggest that the SFA learning can extract useful local motion features and further improve the performance of action recognition.

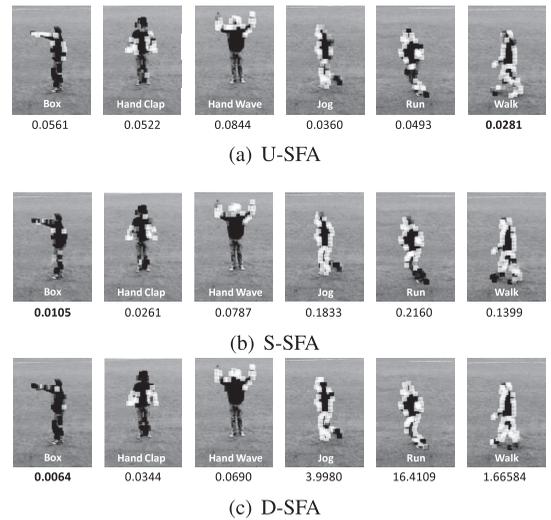

Fig. 12. The squared derivatives of the cuboids transformed by the learned slow feature functions. The value under each figure is the corresponding average squared derivative over all cuboids.

## 6.2 Control Experiment 2: The Characteristics of the Learned Slow Feature Functions

In the experiment, we study the characteristics of the learned slow feature functions. We performed four kinds of SFA learning strategies, i.e., U-SFA, S-SFA, D-SFA, and SD-SFA, from 18,000 training cuboids that were collected from the original action sequences of scenario $S1$ in the KTH data set. Fig. 11 shows the optimal excitatory stimulus and the optimal inhibitory stimulus of the first six slow feature functions by using the method in [46]. For D-SFA and SD-SFA, the tradeoff parameter $\gamma$ is set as 0.2. The optimal stimulus includes three patches because the input vector at each time is the concatenation of three successive patches.

The figure shows that each learned slow feature function can be considered as a filter-like spatiotemporal receptive field. Since the label information of action category and the spatial information are incorporated in the SFA learning, the stripes in the receptive field of D-SFA and SD-SFA are richer. The finding suggests that the slow feature functions learned with label and spatial information can represent finer details in human motion and are more useful for action classification than those functions learned without the label or spatial information.

To further understand the characteristics of the learned functions, we chose typical functions learned by U-SFA, S-SFA, and D-SFA. We observe the responses of these functions to different actions. The functions are the fifth functions learned by the U-SFA, S-SFA, and D-SFA learning strategies, respectively. The slow feature functions learned by S-SFA and D-SFA belong to the action of *boxing*. After the transformation of slow feature functions, the squared derivatives of the cuboids in a certain frame of different actions are shown in Fig. 12. The magnitudes of the squared derivatives are represented by different gray levels. The darker (brighter) the gray level is, the smaller (larger) the magnitude will be. A statistical index—the average squared derivative over all cuboids—is computed to show the statistical characteristic of the learned function (the value under each figure). We find the U-SFA slow

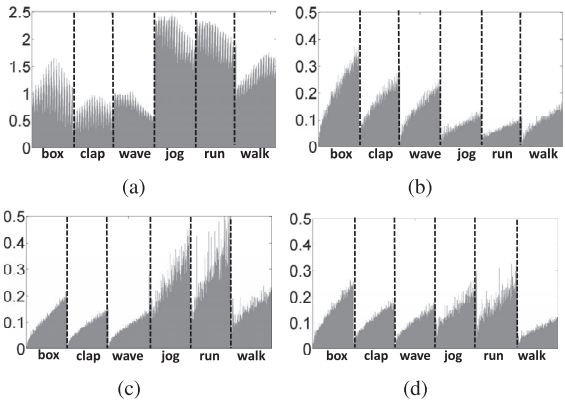

Fig. 13. The average accumulated squared derivatives features of the original cuboids and the cuboids transformed by the U-SFA slow feature functions. (a) The ASD features of the original cuboids. (b) The ASD feature of the cuboids transformed by the U-SFA functions which are learned from the training cuboids of all action categories. (c) The functions are learned from the cuboids without *jogging*, *running*, and *walking*. (d) The functions are learned from the cuboids without *jogging* and *running*.

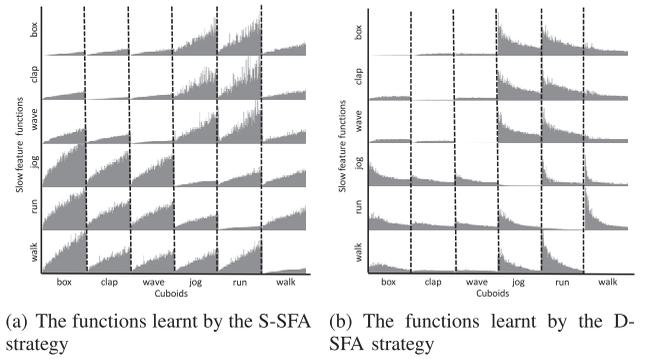

(a) The functions learnt by the S-SFA strategy

(b) The functions learnt by the D-SFA strategy

Fig. 14. The ASD features of the cuboids transformed by the S-SFA and D-SFA slow feature functions.

feature function preferable for the action of *walking*, whose statistical index is the smallest (denoted by bold font). For the S-SFA slow feature function, the smallest index is achieved by *boxing* because the function is learned from the cuboids of *boxing*. For the D-SFA slow feature function, the selectivity of the chosen function to *boxing* is much more remarkable, i.e., the differences between the indices of *boxing* and other actions are much larger. The superior selectivity suggests the feature of D-SFA is more suitable than that of S-SFA for action recognition.

To give an overall evaluation of the functions learned by U-SFA, S-SFA, and D-SFA, 6,000 cuboids (1,000 cuboids per action) were randomly selected and transformed by the first 200 slow feature functions (for S-SFA/D-SFA, 200 slow feature functions per action). Then, ASD features were computed and averaged by the number of cuboids. To demonstrate the capability of SFA to slow down the input signal, an ASD-like feature is also computed with the original cuboid representation, wherein the feature is a 256 ($16 \times 16$) dimensional vector. In Figs. 13 and 14, the feature values belonging to different actions are separated by the dashed lines.

Fig. 13a presents the ASD features of the original cuboid. Fig. 13b presents those of the U-SFA slow feature functions. Compared with Fig. 13a, the transformed cuboids are much slower (indicated by the scale of y-axis). Thus, SFA slows down the input signals. Because of the normalization of the action sequences and the constraints in the SFA learning, both the pixel value and the responses of the learned slow feature functions fall into the same scale (zero mean with unit variance), the comparison is feasible. Fig. 13b also shows that the variations of *jogging*, *running*, and *walking* are smaller than other actions after the U-SFA transformation. This observation suggests that these three actions share many similar motion patterns and thus the learned slow feature functions prefers those shared patterns for slower outputs over all training cuboids. To validate this hypothesis, we train slow feature functions by U-SFA on the cuboid set without *jogging*, *running*, and *walking*. The ASD features are shown in Fig. 13c. Compared with Fig. 13b, which shows the ASD features by using all cuboids in training slow feature functions, we find that without the cuboids of *jogging*, *running*, and *walking*, the feature values of *jogging*, *running*, and *walking* turn larger, whereas the features of the other three actions turn smaller. That indicates the learned slow feature functions mainly encode the motion patterns of the other three actions. Afterward, we add the cuboids of *walking* for SFA learning. As shown in Fig. 13d, we find that the ASD features of *walking*, *jogging*, and *running* jointly turn to smaller. This validates that these three actions share many similar motion patterns.

U-SFA does not consider the supervised information for subsequent recognition and thus we propose S-SFA and D-SFA to learn discriminative slow feature functions. Fig. 14 confirms and compares the effectiveness of S-SFA and D-SFA for action representation. The figure consists of two subplots and each subplot contains $6 \times 6$ cells. The $(i,j)$th cell presents the ASD feature of the $i$th action's training cuboids computed by the $j$th action's slow feature functions. The left subplot shows the feature matrix of S-SFA and the right one is that of D-SFA. The slow feature functions learned by S-SFA and D-SFA are selective to the intraclass cuboids. Or equivalently, the S-SFA and D-SFA slow feature functions for an action make cuboids of the same action as slow as possible, while making cuboids of the other actions vary quickly. Therefore, the relative feature difference between the intraclass cuboids and the interclass cuboids can measure the selectivity of the learned slow feature functions. In contrast to S-SFA, D-SFA considers the interclass information, so the selectivity of D-SFA is stronger than that of S-SFA. To quantify this comparison, we sum the features over all slow feature functions in each cell, and then calculate the ratio between the interclass summed features and the intraclass summed features. Table 1 shows the relative ratios. In each row, the smallest ratio (denoted by bold font) besides the diagonal value is used to measure the selectivity of the slow feature functions. The larger the value is, the stronger the selectivity will be. The average selectivity of the S-SFA slow feature functions is $(1.89 + 3.26 + 3.12 + 2.61 + 3.95 + 4.15)/6 = 3.16$, that of D-SFA is $(2.48 + 5.64 + 4.65 + 5.9 + 6.93 + 5.57)/6 = 5.195$. This confirms that the selectivity of the slow feature functions learned by D-SFA is stronger than the S-SFA functions.

TABLE 1
Quantitative Evaluation of the ASD Features
(S-SFA versus D-SFA)

|       | box   | clap | wave | jog   | run   | walk  |
|-------|-------|------|------|-------|-------|-------|
| $F_b$ | 1     | **1.89** | 1.92 | 7.78  | 8.74  | 3.44  |
| $F_c$ | 4.79  | 1    | **3.26** | 15.1  | 19.18 | 5.61  |
| $F_w$ | 5.01  | **3.12** | 1    | 9.19  | 10.54 | 4.31  |
| $F_j$ | 7.99  | 5.58 | 4.99 | 1     | **2.61** | 2.62  |
| $F_r$ | 10.40 | 7.43 | 6.85 | **3.95** | 1     | 5.17  |
| $F_w$ | 7.87  | 4.8  | 4.25 | **4.15** | 5.21  | 1     |
|       | box   | clap | wave | jog   | run   | walk  |
| $F_b$ | 1     | **2.48** | 2.70 | 25.57 | 25.92 | 10.54 |
| $F_c$ | 8.26  | 1    | **5.64** | 37.38 | 38.94 | 14.89 |
| $F_w$ | 7.9   | **4.65** | 1    | 33.66 | 35.72 | 11.23 |
| $F_j$ | 14.99 | 8.84 | 8.39 | 1     | 8.36  | **5.90** |
| $F_r$ | 11.73 | 7.13 | **6.93** | 9.85  | 1     | 12.7  |
| $F_w$ | 13.86 | 5.66 | **5.57** | 16.63 | 26.7  | 1     |

TABLE 2
Comparison between the SFA-Based Methods and
Previous Methods on the KTH Data Set

| Methods | Test. Str. | Accuracy | Years |
|---------|-----------|----------|-------|
| Schuldt. [26] | RSD | 71.72% | 2004 |
| Dollar [16] | LOO | 81.17% | 2005 |
| Niebles [12] | LOO | 83.33% | 2006 |
| JHuang [32] | RSD | 91.7% | 2007 |
| Schindler [33] | RSD | 92.7% | 2008 |
| Laptev [23] | RSD | 91.8% | 2008 |
| Zhang [27] | RSD | 91.33% | 2008 |
| Liu [29] | LOO | 94.16% | 2008 |
| Wang [28] | LOO | 91.2% | 2009 |
| Bregonzio [34] | LOO | 93.17% | 2009 |
| U-SFA | RSD | 84.67% | |
| S-SFA | RSD | 88.83% | |
| D-SFA | RSD | 91.17% | |
| SD-SFA | RSD | 93.50% | |

### 6.3 Experimental Results on the KTH Data Set

In this section, we report the experimental results on the KTH data set [26]. Consistent with the experiment setting used in [12], [16], [23], [26], [29], we trained and tested the proposed method on the entire data set, in which videos of four scenarios were mixed together. We split the data set into a training part with 16 randomly selected subjects and a test part with the remaining nine subjects. Then, we calculated the average performance over five random splits. Here, we adopted the frame difference as the input. Eighteen thousand local cuboids (3,000 cuboids for each category) were sampled to learn slow feature functions. For D-SFA and SD-SFA, the tradeoff parameter $\gamma$ is 0.2.

Fig. 15 presents the confusion matrices of the classification on the KTH data set by different SFA learning strategies. The column of the confusion matrix represents the instances to be classified, while each row represents the corresponding classification results. The main confusion occurs between *jogging* and *running*. To distinguish the two actions is very challenging because the two actions performed by some subjects are very similar.

Table 2 compares the average recognition ratios of the SFA-based methods with the state-of-the-art results obtained by using the Leave-One-Out (LOO) testing strategy and the Random-Split-Data (RSD) testing strategy. Our results were obtained based on RSD. S-SFA and D-SFA learning strategies perform comparably with the state-of-art methods. SD-SFA learning strategy achieves the highest recognition accuracy among all methods with the RSD testing strategy.

### 6.4 Experimental Results on Weizmann Data Set

On the Weizmann data set [13], we performed the RSD test strategy, where the actions of six subjects were randomly selected to extract slow features, and the remaining three subjects for test. For each action category, 2,000 cuboids were collected to learn slow feature functions based on four learning strategies (U-SFA, S-SFA, D-SFA, and SD-SFA). The cuboid size is $16 \times 16 \times 7$.

Fig. 16 presents two ASD feature examples of *jack* and *run*, where the slow feature functions are learned by D-SFA. For each ASD feature, there are 10 groups of subfeatures corresponding to the 10 action categories. As shown in the top part, the different groups are separated by the dashed lines. Each group corresponds to 200 slow feature functions. Here, one function is chosen as example. The function ID is shown under each subfigure. The squared derivatives of all transformed cuboids are shown in the subfigures. The magnitudes are represented by different gray levels. The darker (lighter) the gray level is, the smaller (larger) the magnitude will be. As shown in the figure, the cuboids transformed by the intraclass slow feature function (in a black frame) vary slower than those transformed by the interclass functions. Thus, as shown in top part, the feature values of the intraclass functions are much smaller than others.

Fig. 17 presents the confusion matrices of different SFA strategies. Table 3 compares the proposed approach with the state-of-the-art methods. The table shows that SD-SFA performs better than existing state-of-the-art methods.

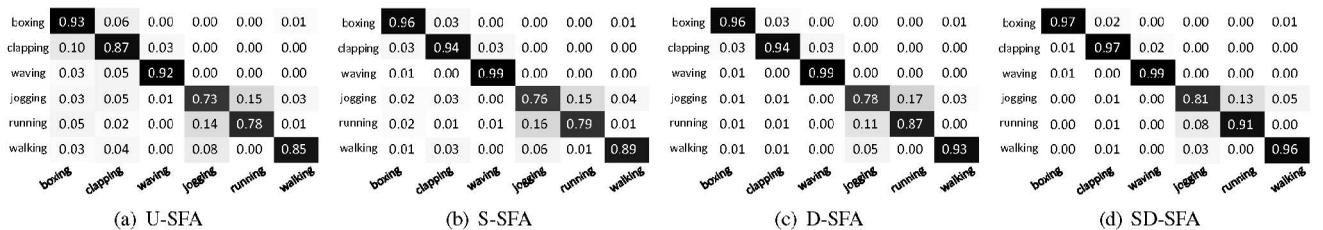

(a) U-SFA　　(b) S-SFA　　(c) D-SFA　　(d) SD-SFA

Fig. 15. Confusion matrices of the classification on the KTH data set obtained by different SFA learning strategies.

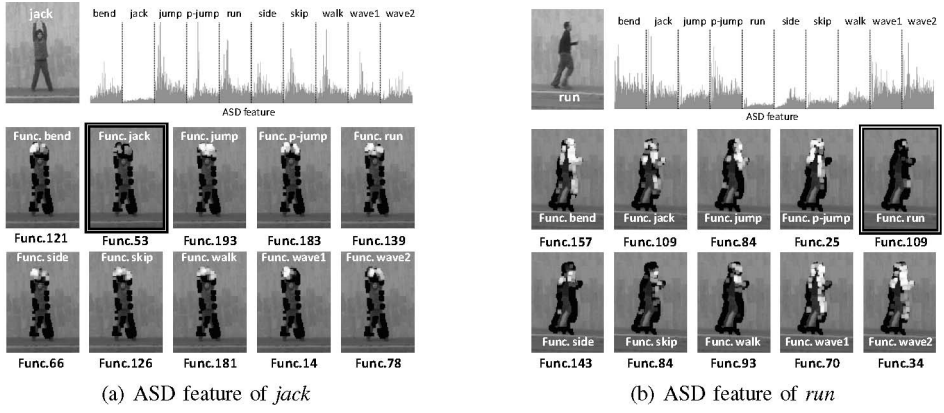

(a) ASD feature of *jack*   (b) ASD feature of *run*

Fig. 16. Examples of the ASD features on the Weizmann data set. The collected cuboids are transformed by 10 sets of slow feature functions learned by the D-SFA. For each set of slow feature functions, one function is selected as an example. The squared derivatives of the transformed cuboids are presented, where the magnitudes are represented by different gray levels.

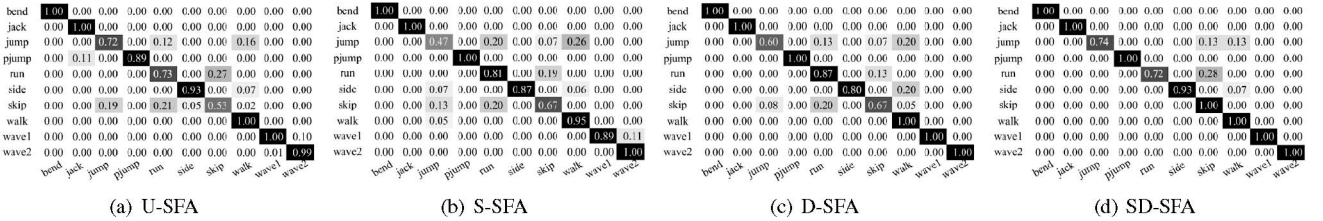

(a) U-SFA   (b) S-SFA   (c) D-SFA   (d) SD-SFA

Fig. 17. Confusion matrices of the classification on the Weizmann data set by different SFA learning strategies.

## 6.5 Experimental Results on CASIA Interaction Data Set

Experiments on both the KTH and the Weizmann data sets already validate the effectiveness of the SFA-based methods for single person action recognition. To demonstrate the advantages of the SFA-based methods, we further conduct experiments on recognizing multiperson interactions in CASIA data set [43], including *meet*, *fight*, and *rob*.

For SFA learning, 10,000 cuboids per action were collected by random sampling in motion regions obtained by frame differencing. To reduce the effects of the background noises (e.g., little camera jitter), only the cuboids changing faster than a threshold are chosen for SFA learning and feature computation. The cuboid size is $8 \times 8 \times 7$, and $\Delta t = 3$. We perform D-SFA to obtain the first 50 slow feature functions for each action. Then, the ASD feature is computed for each frame (note that each feature vector corresponds to the information in seven successive frames). Finally, a linear multiclass SVM [48] is adopted for classification at frame level, where one class label is obtained for every frame and the classification accuracy is computed with (17).

The bag-of-words method was performed for comparison. Each cuboid was represented as a vector by concatenating the brightness gradients in the $x$, $y$, and $t$ directions. The gradient descriptor (GD) proved to be effective for action recognition in [12], [16]. Then, a k-means algorithm was performed to construct the codebook, where the size of the codebook was 150 (50 per action), which was equal to the dimensionality of the ASD feature with D-SFA. Finally, the histogram was formed as a feature vector for the SVM classifier.

Fig. 18 shows the confusion matrices of the fivefold cross validation of the D-SFA + ASD method and the GD + BoW method, respectively. The average performance of our method reaches 97.33 percent, while the performance of bag-of-words model is 73 percent. The experimental results suggest the advantage of the SFA-based method for the complex multiperson activities.

## 6.6 Experimental Results on UT Interaction Data Set

On the UT-Interaction data set, there are 120 action segments of six interaction categories, i.e., *hug*, *kick*, *point*, *punch*, *push*,

TABLE 3
Comparison between the SFA-Based Methods and Previous Methods on the Weizmann Data Set

| Methods | Num. | Test. Str. | Accuracy | Years |
|---|---|---|---|---|
| Niebles [12] | 10 | LOO | 72.8% | 2006 |
| Ali [31] | 9 | LOO | 92.6% | 2007 |
| JHuang [32] | 9 | RSD | 98.8% | 2007 |
| Schindler [33] | 9 | LOO | 100% | 2008 |
| Liu [30] | 10 | LOO | 90.4% | 2008 |
| Klaser [25] | 10 | LOO | 90.5% | 2008 |
| Wang [28] | 9 | LOO | 100% | 2009 |
| Bregonzio [34] | 10 | LOO | 96.66% | 2009 |
| U-SFA | 10 | RSD | 86.67% | |
| S-SFA | 10 | RSD | 86.40% | |
| D-SFA | 10 | RSD | 89.33% | |
| SD-SFA | 10 | RSD | 93.87% | |

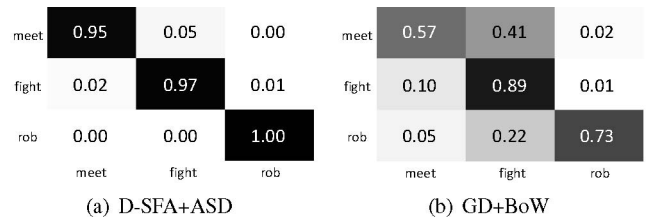

(a) D-SFA+ASD   (b) GD+BoW

Fig. 18. Confusion matrices of multiperson interactions classification: D-SFA + ASD versus GD + BoW.

TABLE 4
Classification Accuracies of Different Methods
on the UT Interaction Data Set, Sets #1 and #2

| Set #1 | Shake | Hug | Kick | Point | Punch | Push | Total |
|---|---|---|---|---|---|---|---|
| Laptev + SVM (best) | 0.5 | 0.8 | 0.7 | 0.8 | 0.6 | 0.7 | 0.683 |
| Cuboid + SVM (best) | 0.8 | 0.9 | 0.9 | 1 | 0.7 | 0.8 | 0.85 |
| Team BIWI | 0.7 | 1 | 1 | 1 | 0.7 | 0.9 | 0.88 |
| **DSFA+ASD+SVM (a)** | 1 | 0.8 | 1 | 1 | 1 | 0.9 | 0.95 |
| **D-SFA+ASD+SVM+Voting (b)** | 1 | 1 | 1 | 1 | 0.8 | 1 | 0.967 |
| **D-SFA+ASD+SVM+Voting (c)** | 1 | 1 | 1 | 1 | 0.9 | 1 | 0.983 |
| Set #2 | Shake | Hug | Kick | Point | Punch | Push | Total |
| Laptev+SVM (best) | 0.5 | 0.7 | 0.8 | 0.9 | 0.5 | 0.5 | 0.65 |
| Cuboid+SVM (best) | 0.8 | 0.8 | 0.6 | 0.9 | 0.7 | 0.4 | 0.7 |
| Team BIWI | 0.5 | 0.9 | 1 | 1 | 0.8 | 0.4 | 0.77 |
| **DSFA+ASD+SVM (a)** | 1 | 0.8 | 1 | 0.9 | 0.8 | 0.9 | 0.9 |
| **D-SFA+ASD+SVM+Voting (b)** | 1 | 1 | 1 | 1 | 1 | 1 | 1 |
| **D-SFA+ASD+SVM+Voting (c)** | 1 | 1 | 1 | 1 | 1 | 1 | 1 |

(a) The feature is computed per segment, (b): per frame, (c): sliding window (*window* size = 7, *step* length = 3).

and *hand-shake*. First, all segments are normalized in terms of [49] so that the main actor of the interaction (e.g., the person punching the other) always stands on the left side. Then, the interest point detector in [16] is adopted to locate the cuboids in action segments, which proved to achieve efficient performance on the UT data set [49]. The cuboid size is $15 \times 15 \times 7$. For D-SFA learning, 5,000 cuboids per category were collected and the first 200 slow feature functions were obtained for the ASD feature computation. Here, the ASD feature was computed by three strategies:

1. For each action segment, one ASD feature was calculated by accumulating the squared derivatives of all transformed cuboids in the segment.
2. For each frame in one action segment, one ASD feature was calculated by accumulating the squared derivatives of all transformed cuboids in the frame.
3. For one action segment, ASD features were calculated by accumulating the squared derivatives of all transformed cuboids in a sliding window ($window\ size = 7$ and $step\ length = 3$).

For the last two feature computing strategies 2 and 3 by which one class label can be obtained per frame or every seven frames, the majority voting rule is adopted to determine the label of a whole segment. In evaluation, there are two test sets, namely, Set #1 and Set #2. Each set contains 60 segments (10 per category). In each set, 10-fold leave-one-out cross validation was performed in terms of the guidance in the contest SDHA. At each round, six interaction segments of one pair of participants were used for the test, and the other 54 segments for the training. Finally, the correct labels in 60 tests are counted to compute the classification accuracies.

The experimental results are presented in Table 4, where the results of "Laptev + SVM (best)," "Cuboid + SVM (best)," and "Team BIWI" are also taken from [49] for comparison. From the tables, the proposed method achieves much higher accuracies than other baseline methods. Especially, the two D-SFA + voting methods can correctly classify all test samples in Set #2.

### 6.7 Discussion

As shown in the experimental results on both the CASIA and the UT data sets, the D-SFA + ASD feature achieves much higher performance than other baseline methods. The distinct performance of the D-SFA + ASD feature in multiperson activities can be explained as follows:

- BoW performs well when: The distribution of local cuboids in an action sequence should be very similar to (very different from) those of other sequences sampled from the same action category (the different action categories) or, briefly, the intraclass variance is large and the interclass variance is small. In general, the intraclass variance of multiperson interactions is much larger than that of single person actions. Therefore, BoW performs well for recognizing single person actions but fails to obtain high performance for recognizing multiperson interactions. For example, in the "boxing" of single person action in KTH data set, the arms of an action subject move back-and-forth alternatively. And in the action of "waving," the subject moves both arms up and down. These two motion patterns are much simpler than that in the "fighting" of multiperson interactions where both motion directions and magnitudes of two participants are varying irregularly. Thus, the BoW feature with gradient-based descriptor (GD + BoW) is ineffective to extract the invariant features in variant local cuboids, which end up with low discriminative capability.
- In contrast with the GD + BoW, SFA indeed extracts slow variant motion patterns hidden in local cuboids as shown in the control experiment 1. In addition, the proposed ASD feature can effectively encode the inherent slow variant motion patterns for action recognition. Thus, SFA with ASD captures the invariance in variant local cuboids. By further considering the discriminative information between different actions, D-SFA functions have strong selectivity to particular actions (shown in the control experiment 2). Therefore, the discriminative capability of the D-SFA + ASD is strong.

We demonstrated the above explanations by using a control experiment, where 100 snippets (each snippet contains seven successive frames) per action category were randomly sampled from the KTH data set, the CASIA data set, and the UT data set (Set #1), respectively. Then, the local cuboids on motion boundaries were obtained by using the method introduced in Section 4. The sizes of cuboids are $16 \times 16 \times 7$, $8 \times 8 \times 7$, and $16 \times 16 \times 7$, respectively. Subsequently, for each cuboid set, the same subset was sampled randomly for both codebook and D-SFA learning. By k-means clustering, 300 codes were obtained for each data set, where each cuboid was described by the gradient-based descriptor in [12], [16]. And 300 D-SFA functions per data set were also learned for computing the ASD feature. Thus, for each snippet in the three sets, one 300-dimension GD + BoW feature and one 300-dimension D-SFA + ASD feature were obtained. Finally, the Fisher score [38] was calculated for each dimension in the features, where the average Fisher scores over 300 dimensions were obtained to measure the discriminative capabilities of the GD + BoW and the D-SFA + ASD features on the three data sets. The results reported in Table 5 indicate that the discriminative capabilities of GD + BoW in the CASIA and the UT data sets are much lower than that in the KTH data set, while those of D-SFA + ASD in the three data sets are comparable. Therefore, D-SFA + ASD significantly outperforms GD + BoW in the case of multiperson activities.

TABLE 5
Comparison of Average Fisher Scores of the D-SFA + ASD Feature and the GD + BoW Feature on Different Data Sets

|  | KTH | CASIA | UT |
|---|---|---|---|
| GD+BoW | 0.15 | 0.03 | 0.02 |
| D-SFA+ASD | 0.20 | 0.22 | 0.17 |

## 7 CONCLUSION

In this paper, we presented four slow feature analysis-based methods for recognizing human actions. The original unsupervised SFA algorithm is extended with different learning strategies, and thus the learned slow feature functions can encode discriminative information for the subsequent recognition. Further, the Accumulated Squared Derivative feature is proposed to characterize the statistical distribution of slow features in an action sequence. To validate the effectiveness of the proposed approach, extensive experiments have been conducted. In particular, there are 1) two sets of control experiments on subsets of the KTH data set, which suggest that SFA can extract effective motion patterns and benefit the performance of action recognition, 2) two sets of experiments on the KTH and Weizmann human action databases to demonstrate that SFA is competitive to the state-of-the-art methods, and 3) two sets of experiments on two multiperson interaction data sets show that the SFA has good potential to recognize complex multiperson activities.

In current experiments, we selected the slowest functions for action representation and the number of functions is determined empirically. In the future, we will investigate strategies to automatically determine the number of the slow feature functions. Furthermore, human motions are composed of several motion parts, and thus the configuration (spatiotemporal structure) between different parts is important to characterize an action. In this paper, the cuboids transformed by the slow feature functions are organized by a simple accumulation of squared derivative. It would be useful and challenging to recovery complex spatiotemporal relationships in the cuboids set for understanding complex human motions. In the future, we will develop strategies to bridge the gap between the slow feature functions and the hidden patterns.

## ACKNOWLEDGMENTS

The authors would like to thank the anonymous reviewers and the handling associate editor for their insightful comments and also thank Dr. P. Dollar for his kind supply of the codes for spatiotemporal cuboid detection. This work is supported in part by the Australian ARC discovery project (ARC DP-120103730), the National Natural Science Foundation of China (Grant Nos. 60875021, 60723005), NLPR (2008 NLPRZY-2), National Hi-Tech R&D Program of China (2009AA01Z318), and National Key Technology R&D Program of China (2011BAH11B01).

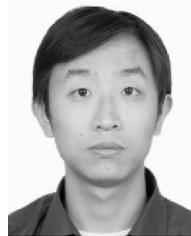

**Zhang Zhang** received the BS degree in computer science and technology from Hebei University of Technology, Tianjin, China, in 2002, and the PhD degree in pattern recognition and intelligent systems from the National Laboratory of Pattern Recognition, Institute of Automation, Chinese Academy of Sciences, Beijing, China in 2008. Currently, he is an assistant professor at the National Laboratory of Pattern Recognition, Institute of Automation, Chinese Academy of Sciences. His research interests include activity recognition, video surveillance, and time series analysis. He has published a number of papers at top venues including the *IEEE Transactions on Pattern Analysis and Machine Intelligence*, CVPR, and ECCV. He is a member of the IEEE.

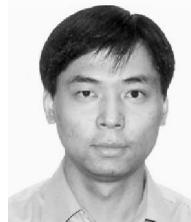

**Dacheng Tao** is a professor of computer science with the Centre for Quantum Computation and Information Systems and the Faculty of Engineering and Information Technology at the University of Technology, Sydney, Australia. He mainly applies statistics and mathematics for data analysis problems in data mining, computer vision, machine learning, multimedia, and video surveillance. He has authored and coauthored more than 100 scientific articles published in top venues including the *IEEE Transactions on Pattern Analysis and Machine Intelligence*, *IEEE Transactions on Knowledge and Data Engineering*, *IEEE Transactions on Image Processing*, NIPS, ICML, UAI, AISTATS, ICDM, IJCAI, AAAI, CVPR, ECCV; *ACM Transactions on Knowledge Discovery from Data*, Multimedia, and KDD, with the best theory/algorithm paper runner up award in IEEE ICDM' 07. He is a member of the IEEE and the IEEE Computer Society.

▷ **For more information on this or any other computing topic, please visit our Digital Library at** www.computer.org/publications/dlib.